\documentclass{article}

\PassOptionsToPackage{numbers}{natbib}

\newcommand\type{preprint}
\usepackage[\type]{neurips_2021}

\usepackage[utf8]{inputenc}   % allow utf-8 input
\usepackage[T1]{fontenc}      % use 8-bit T1 fonts
\usepackage[backref=page]{hyperref} % hyperlinks
\usepackage{url}              % simple URL typesetting
\usepackage{booktabs}         % professional-quality tables
\usepackage{amsfonts}         % blackboard math symbols
\usepackage{nicefrac}         % compact symbols for 1/2, etc.
\usepackage{microtype}        % microtypography
\usepackage{xcolor}           % colors
\usepackage[pdftex]{graphicx} % graphics
\usepackage{graphbox}         % graphics alignment
\usepackage[subrefformat=parens,labelformat=parens]{subcaption} % subfigures
\usepackage{tikz}             % shape drawing
\usepackage{wrapfig}          % figure wrapping
\usepackage{ifthen}           % ifthenelse macro
\usepackage{amsmath}
\usepackage{algorithm}
\usepackage{algorithmicx}
\usepackage{algpseudocode}

\newcommand\mailto[1]{\href{mailto:#1}{\tt{#1}}}
\newcommand\citate[1]{\cite{#1}}
\definecolor{tab_color0}{RGB}{31,119,180}
\definecolor{tab_color1}{RGB}{255,127,14}
\definecolor{tab_color2}{RGB}{44,160,44}
\definecolor{tab_color3}{RGB}{214,39,40}
\definecolor{tab_color4}{RGB}{148,103,189}
\definecolor{tab_color5}{RGB}{140,86,75}
\definecolor{tab_color6}{RGB}{227,119,194}
\definecolor{tab_color7}{RGB}{127,127,127}
\definecolor{tab_color8}{RGB}{188,189,34}
\definecolor{tab_color9}{RGB}{23,190,207}
\DeclareRobustCommand\point[1]{\tikz\draw[#1,fill=#1] (0,0) circle (0.5ex);}

\newcommand{\ifsubmission}[2]{\ifthenelse{\equal{\type}{}}{#1}{#2}}
\newcommand{\meansd}[3]{$\ifthenelse{\equal{#3}{1}}{\textbf{#1}}{#1}\pm#2$}
\DeclareMathOperator{\sign}{sign}
\DeclareMathOperator*{\argmax}{argmax}
\DeclareMathOperator*{\argmin}{argmin}
\newcommand{\popn}{POP-\emph{N}}

\title{Visualizing Representations of Adversarially Perturbed Inputs}

\author{
  Daniel Steinberg \\
  Intelligent Systems Program \\
  University of Pittsburgh \\
  \mailto{das178@pitt.edu} \\
  \And
  Paul Munro \\
  School of Computing and Information \\
  University of Pittsburgh \\
  \mailto{pwm@pitt.edu} \\
}

\begin{document}
  
\maketitle

\begin{abstract}

It has been shown that deep learning models are vulnerable to adversarial
attacks. We seek to further understand the consequence of such attacks on the
intermediate activations of neural networks. We present an evaluation metric,
\popn{}, which scores the effectiveness of projecting data to $N$ dimensions
under the context of visualizing representations of adversarially perturbed
inputs. We conduct experiments on CIFAR-10 to compare the POP-2 score of several
dimensionality reduction algorithms across various adversarial attacks. Finally,
we utilize the 2D data corresponding to high POP-2 scores to generate example
visualizations.

Code is available at~\ifsubmission{\url{https://anonymized/for/submission}}%
{\url{https://github.com/dstein64/vrapi}}.

\end{abstract}

\section{Introduction}

Neural network research has progressed for many decades. Advancements in
computing hardware and model training techniques along with a proliferation of
data has led to widespread adoption of deep learning for solving an assortment
of machine learning problems. Along with its success it has been shown that deep
learning models are susceptible to adversarial
attacks~\citate{szegedy_intriguing_2014}, carefully crafted perturbations of
input instances that appear unaltered to humans and cause incorrect model
output.

Existing work seeks to 1)~develop attack
methods~\citate{kurakin_adversarial_2017}, 2)~develop defense
techniques~\citate{madry_towards_2019}, and 3)~provide insights on the
characteristics of adversarial perturbations and their effects on target
networks~\citate{ma_characterizing_2018,xu_interpreting_2019,tabacof_exploring_2016}.
For a broad overview of the field, see~\citate{akhtar_threat_2018}. Our
work---part of the third aforementioned category---is motivated by trying to
understand how adversarial attacks effect change on the hidden layer
representations of their target networks.

Visualizations have been used to help understand high dimensional data. In this
paper, we focus our attention on visualizing representations of adversarial
inputs at the penultimate layer of the target network. Dimensionality reduction
permits high dimensional data to be projected to 2D, providing data for a
scatter plot. The nature of the task presents a challenge to determine how well
the projected data preserves the high dimensional structure.

\paragraph{Our contribution} In Section~\ref{sec:method} we propose the POP-2
score that can be used for evaluating dimensionality reduction algorithms under
the context of visualizing representations of adversarially perturbed images. We
run experiments in Section~\ref{sec:experimental_results} comparing several
algorithms on CIFAR-10 and present examples of generated visualizations.

\section{Preliminaries}

We are interested in $l$-layer feedforward neural networks, which are functions
\mbox{$h: \mathcal{X} \rightarrow \mathcal{Y}$} that map an input $x \in
\mathcal{X}$ to output $\hat{y} \in \mathcal{Y}$ through linear preactivation
functions $f_i$ and nonlinear activation functions $\phi_i$.

\nopagebreak

\[
\hat{y} = h(x) = \phi_l \circ f_l \circ \phi_{l-1} \circ f_{l-1} \circ \ldots
\circ \phi_1 \circ f_1(x)
\]

We focus here on classification problems, where the output of the neural network
is a discrete label. For an input $x$ and its true class label $y$, we denote
$J(x, y)$ as the corresponding loss of a trained neural network. For convenience,
we omit the dependence on model parameters $\theta$.

\subsection{Adversarial Attacks}

Consider an input $x$ that is correctly classified by neural network $h$. For an
untargeted adversarial attack, an adversary seeks to devise a small additive
perturbation $\Delta x$ such that adversarial input $x^{adv} = x + \Delta x$
changes the classifier's output, i.e., $h(x) \neq h(x^{adv})$. Often the $L_p$
norm of $\Delta x$ is constrained to be smaller than some specified value
$\epsilon$. A targeted adversarial attack additionally specifies a desired value
for $h(x^{adv})$.

For example, an untargeted adversarial input $x^{adv}$ can be generated through
a process that seeks to maximize the loss function. This formulation could be
expressed in terms of a constrained optimization objective as follows.

\nopagebreak

\begin{alignat*}{4}
\Delta x = &\argmax_{\delta}  && J(x + \delta, y) \\
           &\text{subject to} && \  \|\delta\|_p \leq \epsilon \\
           &                  && x + \delta \in \mathcal{X}
\end{alignat*}

There are various ways to formulate the task of generating attacks. An exact
computation of $\Delta x$ is often difficult, thereby entailing that an
approximation be employed.

\textbf{Fast Gradient Sign Method~(FGSM)}~\citate{goodfellow_explaining_2015}
generates adversarial inputs with a perturbation in the approximate direction of
the loss function gradient, whereby $\Delta x$ = $\epsilon \cdot \sign(\nabla_x
J(x, y))$. The $\sign$ function alone constrains the perturbation to an
$L_\infty$ norm bound of 1, which is then scaled by $\epsilon$.

\textbf{Basic Iterative Method~(BIM)}~\citate{kurakin_adversarial_2017} performs
multiple iterations of FGSM with each having a separate gradient update. At each
step, $x^{adv}_{t} = x^{adv}_{t-1} + \alpha \cdot \sign(\nabla_x
J(x^{adv}_{t-1}, y))$, with $x^{adv}_0 = x$. The $L_\infty$ norm is bounded by
$\alpha$ on each iteration, and the final $x^{adv}$ can be constrained to an
$\epsilon$-ball of $x$ by clipping $x^{adv}_t$ accordingly at each step.

\textbf{Carlini \& Wagner (CW)}~\citate{carlini_towards_2017} generates
adversarial inputs by using a gradient descent routine to solve $\Delta x =
\argmin_{\delta} (\|\delta\|_p + c \cdot f(x + \delta))$ subject to a box
constraint on $x + \delta$, where $c > 0$ is a suitable constant and $f$ is a
function for which $f(x + \delta) \leq 0$ if and only if the target classifier
is successfully attacked. The box constraint can be satisfied with clipping or a
change of variables. The most effective $f$ of those considered---in the context
of targeted attacks---was found through experimentation. Binary search is used
to choose the smallest value of $c$, a strategy that worked well empirically.

\subsection{Dimensionality Reduction}

Consider a set of $n$ data points in $D$ dimensions represented with matrix $X
\in \mathbb{R}^{n{\times}D}$. For $d < D$, a dimensionality reduction algorithm
finds lower-dimensional $Z \in \mathbb{R}^{n{\times}d}$ that preserves some
notion of the high dimensional dataset's structure (e.g., pairwise distances).
Techniques differ in various ways, including how much weight is given to
preserving local versus global structure.

An out-of-sample data point is one that is not included in $X$. Some
dimensionality reduction algorithms natively support the projection of arbitrary
$D$-dimensional out-of-sample data points to $d$ dimensions. Out-of-sample
extensions have been proposed for adding support to other
models~\citate{bengio_out--sample_2004}.

\textbf{Principal Components Analysis~(PCA)}~\citate{pearson_liii._1901} is a
linear method that finds a lower-dimensional representation of $X$ that
maximizes the variance of the projected data. A dual formulation seeks to
minimize the sum-of-squared projection errors---the distance of each data point
in high dimension to its projection on a lower-dimensional subspace. The
optimization problem is solved by computing eigenvalues and eigenvectors of
$X$'s covariance matrix $\Sigma$, often conducted using singular value
decomposition (SVD). The $d$ eigenvectors corresponding to the highest
eigenvalues are used for transforming $D$-dimensional data to $d$ dimensions.

\textbf{t-Distributed Stochastic Neighbor
  Embedding~(t-SNE)}~\citate{van_der_maaten_visualizing_2008} is a dimensionality
reduction algorithm designed for visualization. A Gaussian distribution is used
for computing normalized pairwise conditional affinity $p_{j|i}$ between points
$i$ and $j$ in high dimensional space using a bandwidth parameter $\sigma_i$
that satisfies a specified perplexity hyperparameter. The asymmetric
conditionals $p_{i|j}$ and $p_{j|i}$ are symmetrized to $p_{ij}=(p_{j|i} +
p_{i|j})/(2n)$. An optimization procedure is used to position points in low
dimension by minimizing the KL-divergence between $p_{ij}$ and low dimensional
normalized pairwise affinity $q_{ij}$. The formulation of $q_{ij}$ differs from
that of $p_{ij}$, as the former is based on a fatter-tailed Student's
$t$-distribution instead of a Gaussian, and is not a function of asymmetric
conditionals like the latter.

\textbf{Uniform Manifold Approximation and Projection
  (UMAP)}~\citate{mcinnes_umap_2020} is a dimensionality reduction technique
grounded in manifold theory and topological data analysis, which gives rise to a
procedure similar to t-SNE. Pairwise conditional affinity $p_{j|i}$ between
points $i$ and $j$ in high dimensional space is calculated as a function of
distance between $i$ and $j$ shifted and scaled by local connectivity
parameters. The asymmetric conditionals $p_{i|j}$ and $p_{j|i}$ are symmetrized
to $p_{ij}=p_{j|i}+p_{i|j}-p_{j|i}p_{i|j}$. An optimization procedure that
incorporates negative sampling is used to position points in low dimension by
minimizing the cross-entropy between $p_{ij}$ and low dimensional pairwise
affinity $q_{ij}$. The $q_{ij}$ affinities are computed directly---not as a
function of asymmetric conditionals like $p_{ij}$---using two hyperparameters
that control the distance between points in the embedding. Notably, $q_{ij}$ is
not normalized across all data, unlike the similar term in t-SNE.

\textbf{Parametric UMAP}~\citate{sainburg_parametric_2021} is a variation of
UMAP whereby a neural network is trained to map high dimensional points to lower
dimension. The embedding loss function is adapted from UMAP such that the
network learns to position points in a manner similar to using UMAP directly.
The formulation of UMAP---using negative sampling and normalization of pairwise
relationships across all embedding points---makes it particularly suitable for a
neural network variation. The trained parametric model could readily be used to
reduce the dimensionality of out-of-sample data that was not used for training.

\section{Method}
\label{sec:method}

{
  \newcommand\width{0.30\linewidth}
  \begin{figure}[t]
  \centering
  \begingroup
  \renewcommand{\arraystretch}{1.5}
  \begin{tabular}{p{\width}p{\width}p{\width}}
    \multicolumn{1}{c}{\textbf{Step 1}} &
    \multicolumn{1}{c}{\textbf{Step 2}} &
    \multicolumn{1}{c}{\textbf{Step 3}} \\
    \includegraphics[align=c,width=\linewidth]%
      {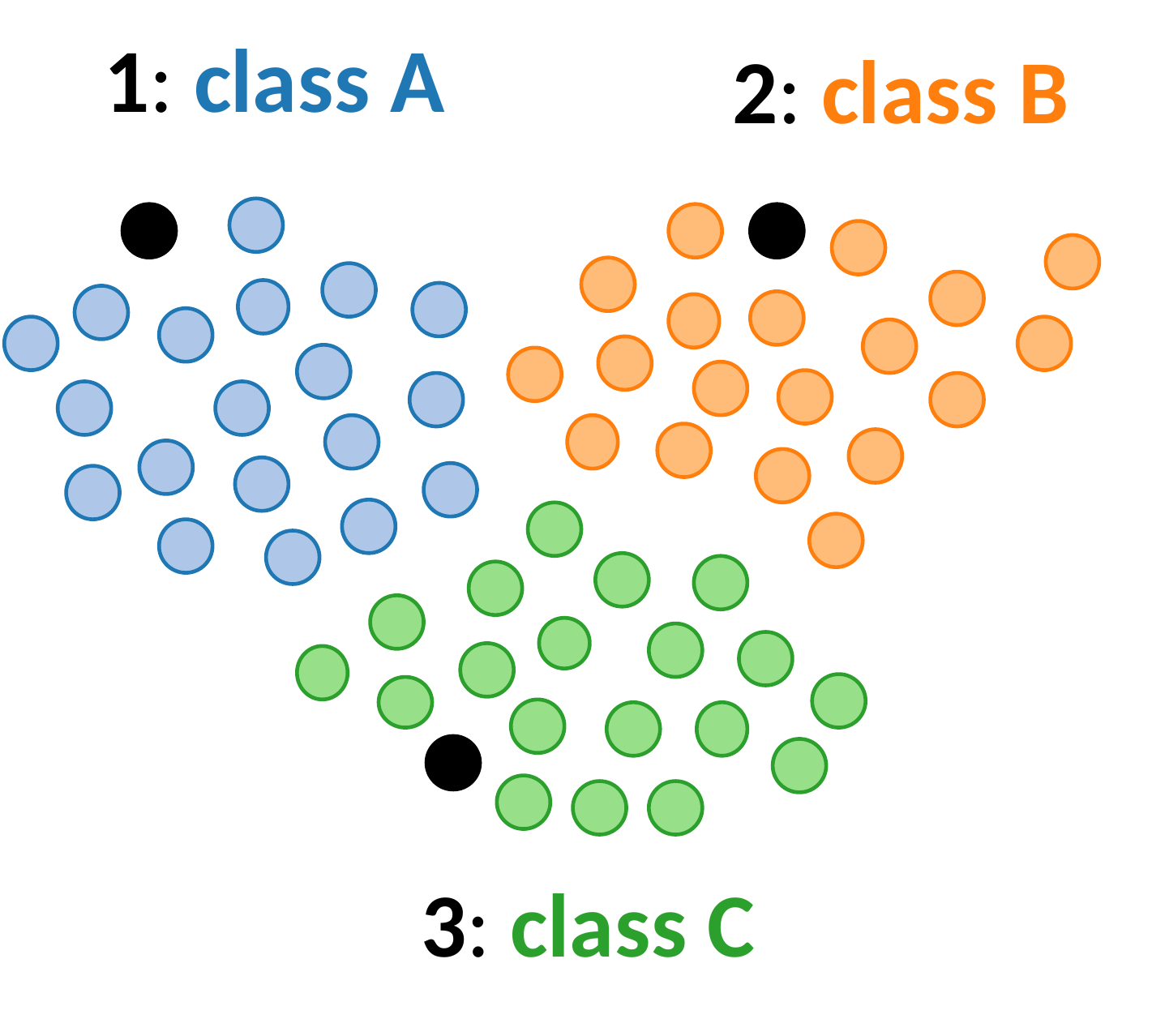} &
    \multicolumn{1}{c}{\includegraphics%
      [align=c,width=\linewidth,height=3.8cm,keepaspectratio]%
      {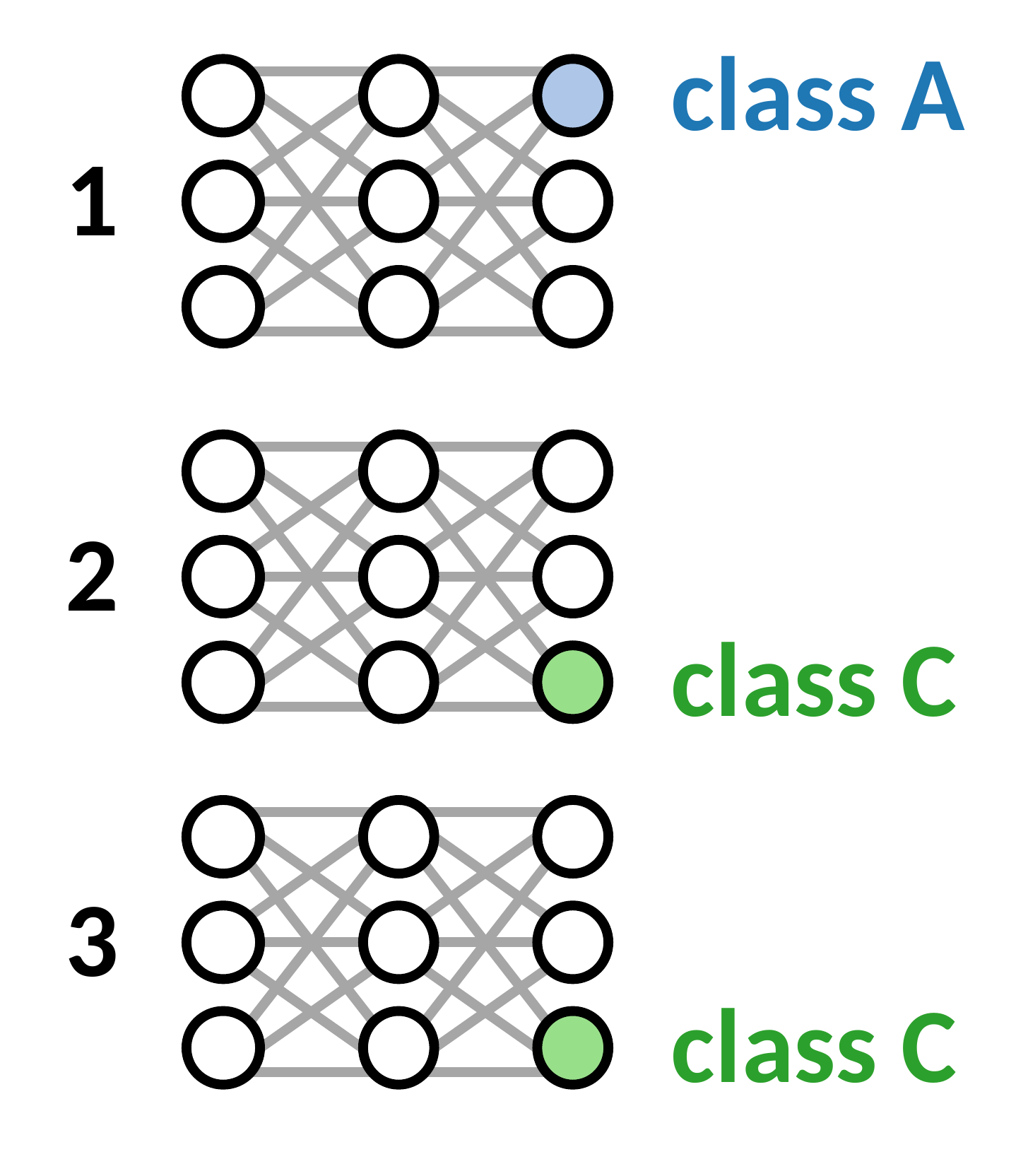}} &
    \includegraphics[align=c,width=\linewidth]%
      {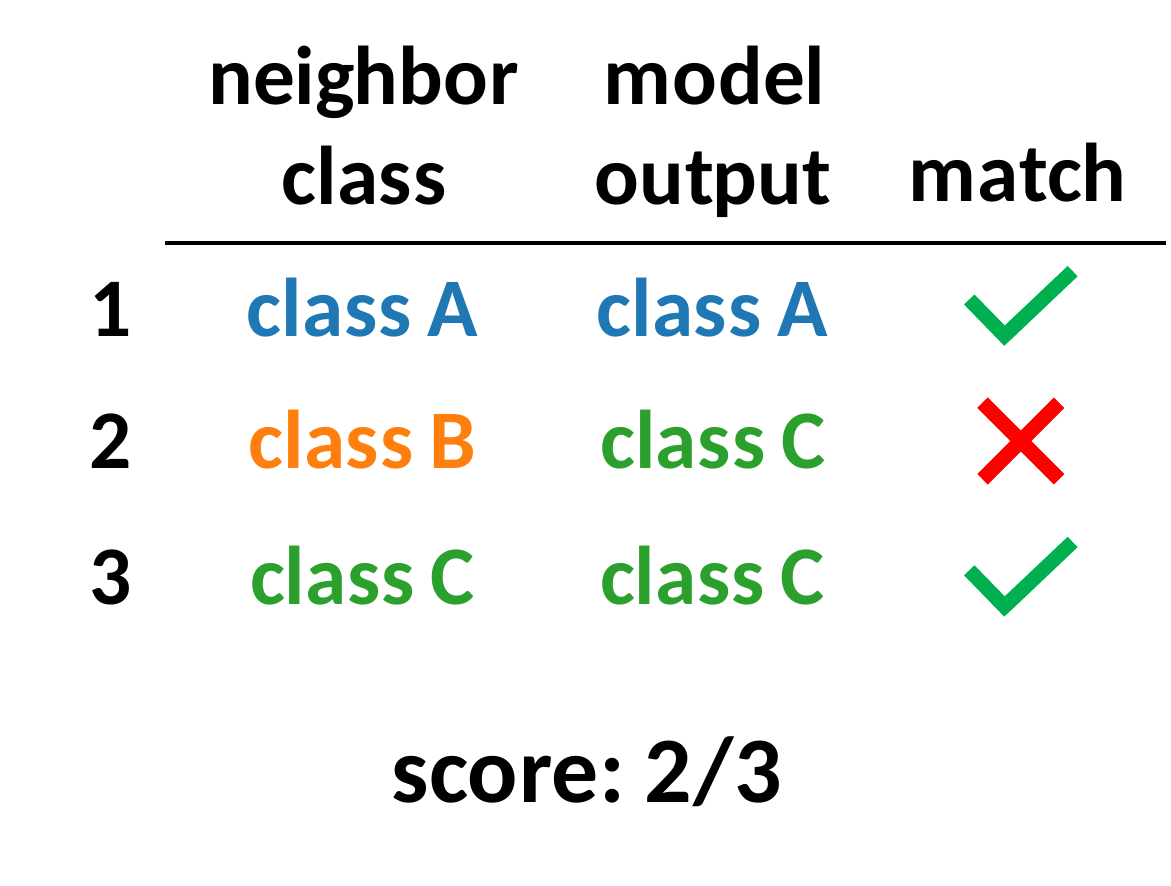} \\
    For each adversarial input---\textbf{\textsf{1}}, \textbf{\textsf{2}}, and
    \textbf{\textsf{3}}---check the class of the closest non-adversarial point in
    the projection. &
    Check the attacked model's predicted class for each adversarial input. &
    Compute \popn{} score based on the consistency of \emph{neighbor class} from
    Step 1 and \emph{model output} from Step 2.
  \end{tabular}
  \endgroup
  \caption{
    Illustration of the \popn{} scoring approach presented in
    Section~\ref{sec:method}.
  }
  \label{fig:popn}
  \end{figure}
}

We're interested in visualizing the penultimate layer representations of
adversarial inputs relative to non-adversarial inputs. At this layer the
representations are typically high dimensional, and thus we need to project the
data to two dimensions for visualization. We propose and define POP-$N$ scoring,
a technique that can be used to evaluate the effectiveness of dimensionality
reduction in the context of visualizing penultimate layer representations of
adversarial images. We then describe how to generate the visualizations.

\subsection{\popn{} Score}

\begin{algorithm}[t]
  \caption{\popn{} Score}
  \begin{algorithmic}[1]
    \Statex \textbf{inputs:}
    \Statex
    \begin{tabular}{ll}
      $Z$       & $N$-dimensional projected representations for correctly
                  classified inputs $X$ \\
      $Y$       & Labels for correctly classified inputs $X$ \\
      $Z^{adv}$ & $N$-dimensional projected representations for $n$
                  adversarially perturbed inputs $X^{adv}$ \\
      $Y^{adv}$ & Predicted labels for $n$ adversarially perturbed inputs
                  $X^{adv}$ \\
    \end{tabular}
    \Function{Score}{}
    \State \textsc{Model} $\gets \textsc{Train}(Z, Y)$
    \label{alg:popn:train}
    \Comment{train nearest neighbor classifier on $Z$ and $Y$}
    \State $c \gets 0$
    \Comment{initialize preserved prediction counter}
    \For{$i \gets 1, \ldots, n$}
    \Comment{loop over adversarial data}
      \If{$Y^{adv}_i = \textsc{Model}(Z^{adv}_i)$}
        \Comment{check if prediction is preserved}
        \State $c \gets c + 1$
        \Comment{increment counter}
      \EndIf
    \EndFor
    \State \Return $c / n$
    \Comment{return preserved prediction rate}
    \EndFunction
  \end{algorithmic}
  \label{alg:popn}
\end{algorithm}

We'd like to generate visualizations from penultimate layer representations of
1)~correctly classified inputs and 2)~adversarial inputs. We need to project the
data to two dimensions for visualization in a scatter plot. To compare
dimensionality reduction algorithms for this task, we devised the \popn{}
score---named as an acronym for \emph{\textbf{p}reserving \textbf{o}utput
  \textbf{p}redictions} in \emph{\textbf{N}} dimensions---for measuring how well
the projected adversarial data preserves the corresponding adversarial output
predictions from the underlying model.

Because the representations under consideration are from the penultimate layer
of a neural network, we utilize class label predictions---which are from the
\emph{neighboring} layer---to evaluate whether the low dimensional embeddings
preserve the \emph{expected} structure in the data. We score a projection by
whether the non-adversarial nearest neighbors of adversarial points have the
same predicted class. The algorithm is illustrated in Figure~\ref{fig:popn} and
described in Algorithm~\ref{alg:popn}.

\subsection{Visualization}

For visualizing adversarially perturbed data, POP-2 can be used to 1)~evaluate
the suitability of a specific dimensionality reduction algorithm and/or
2)~compare scores across an assortment of algorithms. Once an algorithm is
selected, the projected 2D representations can be visualized on a scatter plot.
We follow the approach of \citate{li2020visualizing}, where projected
adversarial representations are visualized alongside projected non-adversarial
representations.

\section{Experimental Results}
\label{sec:experimental_results}

In this section we 1)~compare POP-2 scores across dimensionality reduction
algorithms and adversarial attacks, and 2)~ generate corresponding
visualizations. Hyperparameter selection was performed arbitrarily; we did not
conduct hyperparameter optimization and did not focus on manually tuning
hyperparameters. For the purpose of our experiments, it was sufficient to have
1)~a model that performed well (i.e., not necessarily optimal), 2)~an assortment
of adversarial attacks that worked well (i.e., haven't been tuned for maximal
effectiveness), and 3)~a variety of dimensionality reduction algorithms.

\subsection{Experimental Settings}
\label{subsec:experimental_settings}

Experiments were conducted using the CIFAR-10
dataset~\citate{krizhevsky_learning_2009}, comprised of 60,000 $32{\times}32$
RGB images across 10 classes, split into 50,000 training images and 10,000
testing images. With pixel values scaled by $1/255$ to be between 0 and 1, we
trained an 18-layer, 11,173,962 parameter neural network classifier with a
ResNet-inspired architecture\footnote{We follow the ResNet-18 architecture
  of~\citate{kuangliu_kuangliupytorch-cifar_2021}, which differs in filter counts
  and depth relative to the ResNet-20 architecture used for CIFAR-10 in the
  original ResNet paper~\citate{he_deep_2016}.}. Training was conducted for 100
epochs using Adam~\citate{kingma_adam:_2014} for optimization, with random
horizontal flipping and random crop sampling---on images padded with 4 pixels
per edge. This resulted in 92.76\% accuracy on the test dataset.

\subsubsection{Adversarial Attacks}

\begin{wrapfigure}{R}{0.5\linewidth}
  \vspace{-\intextsep}
  \begin{center}
    {\renewcommand{\arraystretch}{2}
      \begin{tabular}{rcccc}
        & Original & FGSM & BIM & CW \\
        airplane &
        \includegraphics[align=c,width=0.10\linewidth]{%
          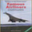} &
        \includegraphics[align=c,width=0.10\linewidth]{%
          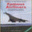} &
        \includegraphics[align=c,width=0.10\linewidth]{%
          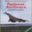} &
        \includegraphics[align=c,width=0.10\linewidth]{%
          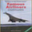} \\
        automobile &
        \includegraphics[align=c,width=0.10\linewidth]{%
          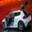} &
        \includegraphics[align=c,width=0.10\linewidth]{%
          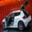} &
        \includegraphics[align=c,width=0.10\linewidth]{%
          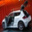} &
        \includegraphics[align=c,width=0.10\linewidth]{%
          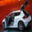} \\
        bird &
        \includegraphics[align=c,width=0.10\linewidth]{%
          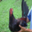} &
        \includegraphics[align=c,width=0.10\linewidth]{%
          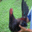} &
        \includegraphics[align=c,width=0.10\linewidth]{%
          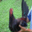} &
        \includegraphics[align=c,width=0.10\linewidth]{%
          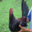} \\
        cat &
        \includegraphics[align=c,width=0.10\linewidth]{%
          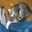} &
        \includegraphics[align=c,width=0.10\linewidth]{%
          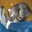} &
        \includegraphics[align=c,width=0.10\linewidth]{%
          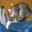} &
        \includegraphics[align=c,width=0.10\linewidth]{%
          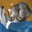} \\
        deer &
        \includegraphics[align=c,width=0.10\linewidth]{%
          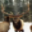} &
        \includegraphics[align=c,width=0.10\linewidth]{%
          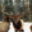} &
        \includegraphics[align=c,width=0.10\linewidth]{%
          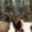} &
        \includegraphics[align=c,width=0.10\linewidth]{%
          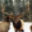} \\
        dog &
        \includegraphics[align=c,width=0.10\linewidth]{%
          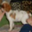} &
        \includegraphics[align=c,width=0.10\linewidth]{%
          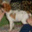} &
        \includegraphics[align=c,width=0.10\linewidth]{%
          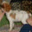} &
        \includegraphics[align=c,width=0.10\linewidth]{%
          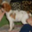} \\
        frog &
        \includegraphics[align=c,width=0.10\linewidth]{%
          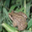} &
        \includegraphics[align=c,width=0.10\linewidth]{%
          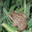} &
        \includegraphics[align=c,width=0.10\linewidth]{%
          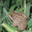} &
        \includegraphics[align=c,width=0.10\linewidth]{%
          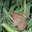} \\
        horse &
        \includegraphics[align=c,width=0.10\linewidth]{%
          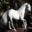} &
        \includegraphics[align=c,width=0.10\linewidth]{%
          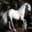} &
        \includegraphics[align=c,width=0.10\linewidth]{%
          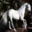} &
        \includegraphics[align=c,width=0.10\linewidth]{%
          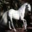} \\
        ship &
        \includegraphics[align=c,width=0.10\linewidth]{%
          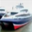} &
        \includegraphics[align=c,width=0.10\linewidth]{%
          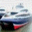} &
        \includegraphics[align=c,width=0.10\linewidth]{%
          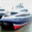} &
        \includegraphics[align=c,width=0.10\linewidth]{%
          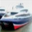} \\
        truck &
        \includegraphics[align=c,width=0.10\linewidth]{%
          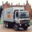} &
        \includegraphics[align=c,width=0.10\linewidth]{%
          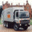} &
        \includegraphics[align=c,width=0.10\linewidth]{%
          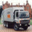} &
        \includegraphics[align=c,width=0.10\linewidth]{%
          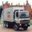}
      \end{tabular}
    }
  \end{center}
  \caption{
    Example CIFAR-10 images after adversarial perturbation. The leftmost
    column shows the original image, followed by three columns corresponding to
    FGSM, BIM, and CW attacks, respectively. Images were chosen as the first of
    their class from the set of---correctly classified without perturbation---test
    images.
  }
  \label{fig:attacked_images}
  %\vspace{-\intextsep}
\end{wrapfigure}

We used the \texttt{cleverhans} library~\citate{papernot2018cleverhans} to
generate untargeted adversarial attacks on the 9,276 correctly classified test
images. Attacks were generated using FGSM, BIM, and CW, with an $L_2$ norm
distance metric used for CW. The perturbed images were clipped between 0 and 1
for all attacks and quantized to 256 discrete values for FGSM and BIM.
Quantization limits the space of perturbed images to those that could be
represented in 24-bit RGB space, but was problematic for CW, where the
perturbations are small enough that quantizing essentially reverts the
attack\footnote{\citate{carlini_towards_2017} addresses this issue through a
  greedy search process that restores attack quality by changing one pixel value
  at a time. We did not utilize that approach for our experiments.}. For FGSM, we
used $\epsilon = 3 / 255$, which corresponds to a maximum perturbation of 3
intensity values for each pixel on the unnormalized data. Model accuracy after
attack---on the 9,276 attacked images---was 17.2\% (i.e., the attack success
rate was 82.8\%). For BIM, we used 10 iterations with $\alpha = 1 / 255$ and the
maximum perturbation magnitude clipped to $\epsilon = 3 / 255$. This corresponds
to a maximum perturbation of 1 unnormalized intensity value per pixel on each
step, ultimately clipped to a maximum perturbation of 3. Accuracy after attack
was 0.5\%. For CW, we used the default parameters---a learning rate of 0.005, 5
binary search steps, and 1,000 maximum iterations. Accuracy after attack was
0.0\%.

Figure~\ref{fig:attacked_images} shows examples of adversarially
perturbed images across the attacks utilized for our experiments. Images were
chosen as the first of their class from the 9,276 correctly classified test
images.

\subsubsection{Dimensionality Reduction}

For the network architecture we used, the penultimate layer representations are
comprised of 512 values formed as the output of a global average pooling
operation. For projecting the corresponding representations to two dimensions,
we use PCA, t-SNE, UMAP, and parametric UMAP. We do not tune hyperparameters,
and we keep the defaults as specified by the libraries we utilized,
\texttt{scikit-learn}~\citate{scikit-learn} and
\texttt{umap-learn}~\citate{mcinnes2018umap-software}. All considered algorithms
include pseudo-randomness, including PCA, for which \emph{randomize SVD} is
automatically chosen as the SVD solver.

We use two approaches for mapping high dimensional representations to 2D. For
the first, model fitting and data transformation are coupled. Non-adversarial
representations $R$ are combined with adversarial representations $R^{adv}$ and
the data is jointly projected to 2D. For the second approach---limited to
dimensionality reduction algorithms that support out-of-sample projection---we
fit models using only non-adversarial representations $R$, producing a
transformation function that is used for projecting both $R$ and $R^{adv}$. A
benefit to utilizing the latter approach is that it more readily permits a
comparison of visualizations across attacks since the non-adversarial
projections are held constant.

\subsection{POP-2 Scores}

We calculate POP-2 scores across the adversarial attacks and dimensionality
reduction algorithms. The results are reported on Table~\ref{table:pop2_scores},
where the values are averaged across 100 trials, each using a different seed for
dimensionality reduction.

\begin{table}[t]
  \caption{
    Average POP-2 scores plus/minus the sample standard deviation, calculated
    across 100 trials that each uses a different seed for 2D dimensionality
    reduction. Rows are sorted in descending order by the mean score across
    attacks. For each attack, the highest POP-2 score is in bold. \emph{OOS} is
    used for labeling the out-of-sample approaches.
  }
  \centering
  \begin{tabular}{lccc}
    \toprule
    & \multicolumn{3}{c}{Adversarial Attack} \\
    \cmidrule(r){2-4}
    \begin{tabular}{l}Dimensionality Reduction \end{tabular}
      & FGSM & BIM & CW \\
    \midrule
    UMAP
      & \meansd{0.8622}{0.0047}{1} & \meansd{0.9950}{0.0004}{1} & \meansd{0.3467}{0.0065}{1} \\
    Parametric UMAP
      & \meansd{0.8589}{0.0114}{0} & \meansd{0.9901}{0.0056}{0} & \meansd{0.3365}{0.0116}{0} \\
    t-SNE
      & \meansd{0.8554}{0.0024}{0} & \meansd{0.9948}{0.0005}{0} & \meansd{0.2144}{0.0027}{0} \\
    OOS parametric UMAP
      & \meansd{0.8188}{0.0105}{0} & \meansd{0.9765}{0.0094}{0} & \meansd{0.2625}{0.0173}{0} \\
    PCA
      & \meansd{0.2903}{0.0000}{0} & \meansd{0.4062}{0.0000}{0} & \meansd{0.1966}{0.0000}{0} \\
    OOS PCA
      & \meansd{0.2891}{0.0000}{0} & \meansd{0.3767}{0.0000}{0} & \meansd{0.2056}{0.0000}{0} \\
    OOS UMAP
      & \meansd{0.1592}{0.0402}{0} & \meansd{0.1383}{0.0345}{0} & \meansd{0.1202}{0.0377}{0} \\
    \bottomrule
  \end{tabular}
  \label{table:pop2_scores}
\end{table}

\subsection{Visualizations}

{
  \newcommand\width{0.400\linewidth}
  \newcommand\padding{0.05\linewidth}
  \begin{figure}[t]
    \centering
    \begin{subfigure}[t]{\width{}}
      \centering
      \includegraphics[width=\linewidth]{assets/fgsm_umap/%
        non_adversarial-crop.png}
      \caption{Non-adversarial}
      \label{subfig:umap_non_adversarial}
    \end{subfigure}
    \hspace{\padding}
    \begin{subfigure}[t]{\width{}}
      \centering
      \includegraphics[width=\linewidth]{assets/fgsm_umap/%
        non_adversarial_and_adversarial_by_class_0_airplane-crop.png}
      \caption{Airplane}
      \label{subfig:umap_airplane}
    \end{subfigure}
    \\[1.0\baselineskip]
    \begin{subfigure}[t]{\width{}}
      \centering
      \includegraphics[width=\linewidth]{assets/fgsm_umap/%
        non_adversarial_and_adversarial_by_class_1_automobile-crop.png}
      \caption{Automobile}
      \label{subfig:umap_automobile}
    \end{subfigure}
    \hspace{\padding}
    \begin{subfigure}[t]{\width{}}
      \centering
      \includegraphics[width=\linewidth]{assets/fgsm_umap/%
        non_adversarial_and_adversarial_by_class_2_bird-crop.png}
      \caption{Bird}
      \label{subfig:umap_bird}
    \end{subfigure}
    \begin{center}
      \begin{tabular}{lllll}
        \point{tab_color0} airplane & \point{tab_color1} automobile &
        \point{tab_color2} bird     & \point{tab_color3} cat        &
        \point{tab_color4} deer \\
        \point{tab_color5} dog      & \point{tab_color6} frog       &
        \point{tab_color7} horse    & \point{tab_color8} ship       &
        \point{tab_color9} truck \\
        \point{black} adversarial
      \end{tabular}
    \end{center}
    \caption{
      Visualizations of CIFAR-10 test data penultimate layer representations,
      projected to two dimensions with UMAP.
      \subref{subfig:umap_non_adversarial} shows only the projected
      representations of non-adversarial data. \subref{subfig:umap_airplane},
      \subref{subfig:umap_automobile}, and \subref{subfig:umap_bird} show
      faded non-adversarial points along with black points corresponding to
      FGSM adversarially perturbed airplanes, automobiles, and birds,
      respectively.
    }
    \label{fig:umap}
  \end{figure}
}

{
  \newcommand\width{0.450\linewidth}
  \newcommand\padding{0.0\linewidth}
  \begin{figure}[t]
    \centering
    \begin{subfigure}[t]{\width{}}
      \centering
      \includegraphics[width=\linewidth]{assets/parametric_umap_oos/%
        non_adversarial-crop.png}
      \caption{Non-adversarial}
      \label{subfig:parametric_umap_oos_non_adversarial}
    \end{subfigure}
    \hspace{\padding}
    \begin{subfigure}[t]{\width{}}
      \centering
      \includegraphics[width=\linewidth]{assets/parametric_umap_oos/%
        fgsm_non_adversarial_and_adversarial_by_class_0_airplane-crop.png}
      \caption{FGSM}
      \label{subfig:parametric_umap_oos_fgsm_airplane}
    \end{subfigure}
    \\[1.0\baselineskip]
    \begin{subfigure}[t]{\width{}}
      \centering
      \includegraphics[width=\linewidth]{assets/parametric_umap_oos/%
        bim_non_adversarial_and_adversarial_by_class_0_airplane-crop.png}
      \caption{BIM}
      \label{subfig:parametric_umap_oos_bim_airplane}
    \end{subfigure}
    \hspace{\padding}
    \begin{subfigure}[t]{\width{}}
      \centering
      \includegraphics[width=\linewidth]{assets/parametric_umap_oos/%
        cw_non_adversarial_and_adversarial_by_class_0_airplane-crop.png}
      \caption{CW}
      \label{subfig:parametric_umap_oos_cw_airplane}
    \end{subfigure}
    \begin{center}
      \begin{tabular}{lllll}
        \point{tab_color0} airplane & \point{tab_color1} automobile &
        \point{tab_color2} bird     & \point{tab_color3} cat        &
        \point{tab_color4} deer \\
        \point{tab_color5} dog      & \point{tab_color6} frog       &
        \point{tab_color7} horse    & \point{tab_color8} ship       &
        \point{tab_color9} truck \\
        \point{black} adversarial
      \end{tabular}
    \end{center}
    \caption{
      Visualizations of CIFAR-10 test data penultimate layer representations,
      projected to two dimensions with out-of-sample parametric UMAP.
      \subref{subfig:parametric_umap_oos_non_adversarial} shows only the
      projected representations of non-adversarial data.
      \subref{subfig:parametric_umap_oos_fgsm_airplane},
      \subref{subfig:parametric_umap_oos_bim_airplane}, and
      \subref{subfig:parametric_umap_oos_cw_airplane} show faded non-adversarial
      points along with black points corresponding to FGSM, BIM, and CW
      adversarially perturbed airplanes, respectively.
    }
    \label{fig:oos_parametric_umap}
  \end{figure}
}

Figure~\ref{fig:umap} shows example visualizations generated using UMAP for
dimensionality reduction, comparing the adversarial projections across the first
three classes. Figure~\ref{fig:oos_parametric_umap} shows example visualizations
generated using out-of-sample parametric UMAP for dimensionality reduction,
comparing adversarial airplane (selected as the dataset's first class)
projections across the three attacks.

\subsection{Hardware and Software}
\label{subsec:hardware_and_software}

All experiments were conducted on a desktop computer with Ubuntu 20.10, using
Python 3.9. The hardware includes an AMD Ryzen 7 2700X CPU, 64GB of memory, and
an NVIDIA TITAN RTX GPU with 24GB of memory. The GPU was utilized for training
the neural network and generating adversarial attacks.

The code used for running the experiments is available
at~\ifsubmission{\url{https://anonymized/for/submission}}%
{\url{https://github.com/dstein64/vrapi}}.

\section{Discussion}
\label{sec:discussion}

\subsection{POP-2 Scores}

As shown in Table~\ref{table:pop2_scores}, the UMAP algorithm corresponded to
the maximum average POP-2 score for each of the attacks considered. The scores
for BIM were highest, which is interesting particularly relative to the lower
scores for FGSM, given the relationship between those algorithms. Notably, the
CW scores are lowest, suggesting there may be issues with using the POP-2 metric
with that attack. We discuss this further in Section~\ref{sec:conclusion}.

The variability across trials was minor, as shown by the reported standard
deviations. Out-of-sample UMAP had the highest variability and also the lowest
scores, making it particularly unsuitable for this problem.

For each adversarial input, the \popn{} score considers the class of the closest
non-adversarial point in the space of representation projections. An alternative
formulation of \popn{} scoring could hypothetically use nearest class centroids
in place of nearest neighbors. In such a scenario, line~\ref{alg:popn:train} of
Algorithm~\ref{alg:popn} would train a nearest centroid classifier instead of a
nearest neighbor classifier. We intentionally avoided this approach because it
would not properly accommodate 1)~projections with multiple clusters per class,
2)~non-spherical class clusters, and 3)~unevenly sized class clusters.

\subsection{Visualizations}

\paragraph{Non-adversarial data} One of the first patterns we noticed was in the
structure of the non-adversarial data. In Figures~\ref{fig:umap}
and~\ref{fig:oos_parametric_umap}, and other plots we inspected, the data are
organized in clusters that are systematically arranged, as has been often found
in visualizations of hidden unit representations. Each class forms its own
cluster, with the set of vehicle clusters on one side and animal clusters on the
other side, joined by the two aerial classes, airplane and bird.

\paragraph{Comparing classes} Figure~\ref{fig:umap} shows the differences in
projected representations of attacked inputs for an assortment of classes. It is
noticeable how each class of attacked image is unique in the general way its
projected representations manifest. For example, many of the points for the
adversarially perturbed airplanes are concentrated in the region connecting
airplane and bird clusters. For the adversarially perturbed automobiles, a large
portion of the projections appear to uniformly fill the truck cluster. The bird
class's projected representations are spread most widely.

\paragraph{Comparing attacks} On Figure~\ref{fig:oos_parametric_umap} we can see
the differences in projected representations for the adversarially perturbed
airplanes across the different attacks. Comparing FGSM and BIM, it appears that
the representations for adversarial inputs are occupying different regions of
each class's representation space. For example, low dimensional representations
for the FGSM adversarially perturbed airplane images occupy the leftmost portion
of the ship cluster, whereas the corresponding points for BIM occupy the
rightmost portion. For CW, it's noticeable how the points for adversarially
perturbed airplane inputs are concentrated slightly outside the airplane
cluster. This may be reflective of the minor magnitude of the CW perturbation
distance. It's also noticeable that a non-negligible amount of adversarial
density occupies a region of the embedding---above the airplane cluster---that
is sparse for non-adversarial data. This was not the case for FGSM and BIM
attacks.

\section{Related Work}

\paragraph{Grand Tour} Work similar to ours is presented
in~\citate{li2020visualizing}, where one of the example use cases for the
proposed Grand Tour visualization technique is in the context of adversarial
examples. They show how Grand Tour can be used to visualize the layer-by-layer
behavior of adversarial image representations for MNIST images. In addition to
showing the dynamics throughout different layers, they also visualize how
representations change as a function of the optimization process used for
generating an adversarial image. Some aspects that distinguish our work from
theirs include 1)~proposing a metric for evaluating visualizations, 2)~using the
CIFAR-10 dataset as opposed to MNIST, 3)~considering an assortment of
dimensionality reduction techniques---both linear and non-linear, in contrast to
Grand Tour, which is ``fundamentally a linear method''---for visualizing
adversarial perturbations, 4) including more points in our visualizations, 5)
focusing on untargeted adversarial attacks instead of targeted attacks, and 6)
considering an assortment of adversarial attacks.

\paragraph{Adversarial attack visualizations} More broadly, there are various
works that use visualizations as a way to better understand adversarial attacks.
\citate{xu_interpreting_2019}~proposes a technique for image-level
interpretability of adversarial examples, whereby class activation maps and
perturbation information are overlaid on original and adversarially attacked
images. \citate{norton_adversarial-playground_2017}~describes
Adversarial-Playground, an interactive web application that shows original
images alongside adversarial counterparts, intended to help users visually
explore the behavior of adversarial attacks and how they impact their targeted
classifiers. In addition to displaying---original and adversarial---images
themselves, visualizations of decision boundaries have been another technique
used to understand classifier vulnerability to adversarial inputs. For example,
\citate{liu_delving_2017}~presents such visualizations---see Figures 3, 4, and 5
in that paper---as part of their research on adversarial example
transferability.

\section{Conclusion}
\label{sec:conclusion}

The POP-2 score can be used to evaluate dimensionality reduction algorithms
under the context of visualizing representations of adversarially perturbed
inputs. We showed how this can be used to compare algorithms and/or check the
performance of a specific algorithm. Experiments indicate that there is a wide
difference in effectiveness across different dimensionality reduction algorithms
and adversarial attacks.

The research leaves open unanswered questions, suggesting various avenues for
future work. The \mbox{POP-2} metric was devised with specific consideration for
representations from the penultimate layer of neural networks. Devising metrics
for earlier layers would permit a quantitative evaluation of dimensionality
reduction techniques on such layers. In the absence of a method for quantifying
performance in earlier layers, the representations could be visualized using the
same dimensionality reduction algorithms that scored high for the penultimate
layer, but there would be more uncertainty regarding the visualization
accuracy\footnote{The term ``accuracy" is used loosely here, as it's not well
  defined in this context.}. Finally, future work could additionally address the
low POP-2 scores for the CW attack, exploring why this occurs and either 1)
devising a metric that would potentially be more suitable than POP-2 for that
attack, or 2) searching for or developing alternative dimensionality reduction
algorithms that would potentially be more applicable for that attack.

\section{Broader Impact}

Visualizing the representations of adversarially perturbed images contributes to
a further understanding of 1)~neural networks and 2)~adversarial attacks and
defenses against neural networks. While this knowledge could conceivably be
utilized by a malicious actor, we believe the potential benefits provided by
this area of research outweigh the potential for misuse.

% The ack environment is automatically not shown for submission (i.e., when
% [preprint] and [final] are omitted).

%\begin{ack}
%\end{ack}

% **************************************
% * References
% **************************************

\clearpage

% 1) Any choice of citation style is acceptable as long as you are consistent.
% 2) It is permissible to reduce the font size to \small (9 point) when
% listing the references.
% 3) Note that the Reference section does not count towards the page limit.

{
\small
\bibliography{paper}
% Built-in standard styles include plain, acm, ieeetr, alpha, abbrv, and siam.
% For others, a corresponding bst file is necessary. Some styles require the
% [numbers] option for natbib if they use numerical citations.
\bibliographystyle{alpha}
}

% Add References link to Contents. This has to be after \biliography, as the
% section is specified with that command.
\addcontentsline{toc}{section}{References}

\end{document}